\documentclass[12pt]{article}

\usepackage{subfig}
\usepackage{graphicx}
\usepackage{hyperref}
\usepackage{breakcites}
\usepackage[square]{natbib}

\title{Barista -- a Graphical Tool for Designing and Training Deep Neural Networks}
\author{	Klemm, S\"oren\	\texttt{s.klemm@wwu.de}$^\star$
	\and
	Scherzinger, Aaron\ \texttt{scherzinger@wwu.de}\footnote{These authors contributed equally to this work.}
	\and
	Drees, Dominik\ \texttt{dominik.drees@wwu.de}
	\and
	Jiang, Xiaoyi\ \texttt{xjiang@wwu.de}\bigbreak \\
	Faculty of Mathematics and Computer Science, \\University of M\"{u}nster, M\"{u}nster, Germany
}

\newcommand{\baristaurl}{\url{https://barista.uni-muenster.de}}

\begin{document}

\maketitle

\begin{abstract}
In recent years, the importance of deep learning has significantly increased in pattern recognition, computer vision, and artificial intelligence research, as well as in industry. 
However, despite the existence of multiple deep learning frameworks, there is a lack of comprehensible and easy-to-use high-level tools for the design, training, and testing of deep neural networks (DNNs).
In this paper, we introduce Barista, an open-source graphical high-level interface for the Caffe deep learning framework. 
While Caffe is one of the most popular frameworks for training DNNs, editing prototext files in order to specify the net architecture and hyper parameters can become a cumbersome and error-prone task.
Instead, Barista offers a fully graphical user interface with a graph-based net topology editor and provides an end-to-end training facility for DNNs, which allows researchers to focus on solving their problems without having to write code, edit text files, or manually parse logged data.
\end{abstract}
       

\section{Introduction}
\label{sec:Introduction}

In recent years, deep learning with neural networks (NNs) has led to major breakthroughs in machine learning and pattern recognition and in many tasks surpassed traditional model-based approaches or other learning methods by a large margin \citep{Schmidhuber15}.
The increasing importance of deep learning in research and industry has led to the development of a broad range of libraries like Caffe, TensorFlow, Theano, and Torch~\citep{Yangqing2014, Abadi2016, Al-Rfou2016, Collobert2011}. 
However, all current deep learning frameworks still require detailed knowledge of the underlying machine learning algorithms and libraries as well as programming experience or tedious editing of text configuration files to set a vast amount of parameters. 
Although there exist graphical interfaces such as NVIDIA's DIGITS\footnote{\url{https://github.com/NVIDIA/DIGITS}}, Intel's Deep Learning SDK\footnote{\url{https://software.intel.com/en-us/deep-learning-sdk}}, Caffe Gui Tool\footnote{\url{https://github.com/Chasvortex/caffe-gui-tool}}, or Expresso~\citep{Dholakiya2015}, none of the products currently available offer a graphical interface for the complete pipeline of deep learning in various applications.

Here, we propose \textit{Barista}, an open-source graphical tool for designing and training deep neural networks. 
Barista uses Caffe as the underlying framework due to its concept of network layers as the basic building blocks of a model, its wide hardware support, and high speed~\citep{Shi2016}. 
Moreover, Caffe is widely used in recent deep learning research~\citep{Chen2017,ScherzingerKlemm17,Yang2017}, provides a wide range of well-established architectures and pre-trained models~\citep{Krizhevsky2012,Ronneberger15} and offers various branches and forks with additional functionality such as Spark\footnote{\url{https://github.com/yahoo/CaffeOnSpark}} or MPI support~\citep{Lee2015}.

\section{System Overview}
\label{sec:System}

\subsection{Integration with Caffe}

Since Barista is designed on top of the Caffe infrastructure, it provides the user with Caffe's available layers, parameters, and solver options. 
However, due to the fast-paced development of deep learning in general and the Caffe framework in particular, the available components may change between Caffe versions. 
Moreover, depending on the application, users may need to switch between different branches of Caffe or specific implementations containing custom layers or solver types. 
Hence, Barista was designed to be capable of supporting all branches that provide a valid \texttt{caffe.proto} file and Python interface.

The user can specify a Caffe version to use. 
Barista will then automatically extract the available layers and parameters from the \texttt{caffe.proto} file. 
This allows to use different Caffe versions as well as branches and forks. 
To allow maximum flexibility, different Caffe versions can be used on a per-project basis.
Furthermore, this simple caffe interface increases portability and compatibility with future Caffe versions without the need to adapt Barista to a particular Caffe implementation.

\subsection{Barista's Project Structure}

Barista stores all information about a NN design in a project folder, comprising model topology, hyper parameters, information about training and validation data, and settings for the graphical user interface including the graphical NN representation.
Training progress and results are managed using \textit{sessions}, i.e., training runs within a project, allowing the user to run and evaluate multiple designs in parallel or switch to different hyper parameter settings or NN topologies.
Session data is stored using native Caffe file formats (i.e., \texttt{*.prototxt} and \texttt{*.caffemodel} as well as Caffe's log files), allowing direct use of (intermediate) results on machines without a Barista installation.
Equally, Barista allows the import of existing Caffe net definitions and solvers as well as learned weights and training snapshots.
Due to this project folder structure, Barista projects are easily transferable between machines.

\subsection{Defining the Neural Network's Topology and (Hyper) Parameters}

\begin{figure*}[tpb]
\centering
\subfloat[]{
\includegraphics[width=0.95\textwidth]{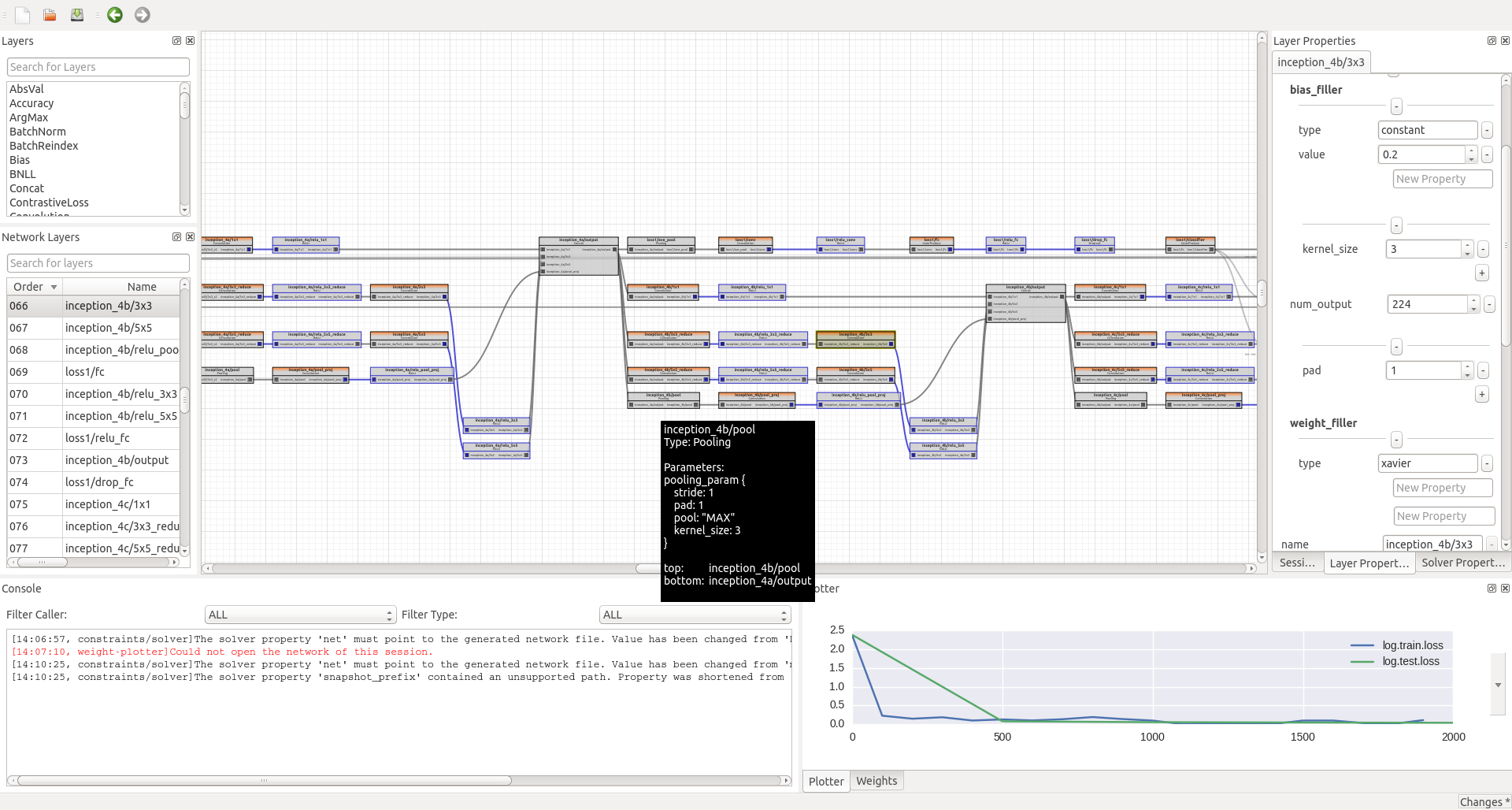}
\label{fig:networkeditor}
}%

\subfloat[]{
\includegraphics[width=0.5\textwidth]{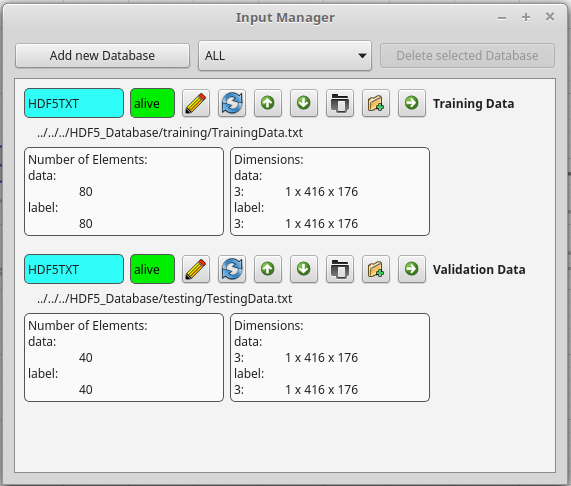}
\label{fig:inputmanager}
}
\label{fig:barista_gui}
\caption{Barista's graphical user interface. (a)~The network editor displays the current network topology and parameters and allows the user to modify the model architecture and parameters or design a new model from scratch. (b)~The input manager allows the user to specify training and validation data and assign it to the NN's data layers.}
\end{figure*}

The central part of Barista's user interface is the \textit{network editor}, containing a graphical representation of the NN (see Fig.~\ref{fig:networkeditor}). 
This representation is generated automatically when importing a NN definition from a prototext file but can also be adjusted by using Drag\&Drop.
Specific layer types are color-coded to improve clarity of a network design.
Data connections like ground truth labels can optionally be hidden to declutter the network graph.
The network editor allows direct manipulation of the network, like adding or deleting layers and editing connections, without the need to type a single line of prototext definition.

Further settings are controlled using different docks, which can be activated and positioned freely by the user.
Parameters of a specific layer can be set using the layer properties dock.
Here, the user is provided with a list of all available parameter groups and parameters for the selected layer.
Depending on the parameter type a list of valid settings is shown, further reducing the need to refer to the Caffe documentation for available options.
Hyper parameters to control training of the NN can be set using a similar interface in the solver properties dock.
A list of all layer types of the selected Caffe branch is available in the layers dock. 
New layers can be added to the network editor using Drag\&Drop.
Editing larger networks is simplified by a text search provided in the network layers dock.

\subsection{Selecting Training and Validation Data}

Barista's \textit{input manager} (see Fig.~\ref{fig:inputmanager}) allows for managing different datasets for training and testing. 
The input manager supports all database types readable by Caffe's data layers. 
When adding data stored in these formats, an overview of the available data (e.g., number of items and dimensionality) is provided.
Moreover, the input manager allows the user to directly assign available databases to different inputs of the NN for training or validation. 

\subsection{Training a Neural Network and Monitoring the Progress}

The \textit{session manager} provides an overview of all sessions in the project and their current progress and allows pausing and resuming training at the click of a button.
Failed or unnecessary training runs can be deleted to free up hard disk memory.
During training, Barista allows to monitor the progress by plotting available Caffe outputs (e.g., training and validation loss or accuracy) in real-time.
The user can visualize several sessions at once to compare parameter sets or NN topologies within the project or import other training log files.
Furthermore, the selected plots can be exported to a CSV file.

The sessions within a project are browsable, i.e., the user can restore the net topology and parameter settings of each session to continue working with this state. 
This allows to evaluate different parameter sets and select the best configuration for further work.

Besides training a model from scratch, Barista also supports the concept of transfer learning, i.e., fine-tuning a pre-trained model using application-specific training data. This can be achieved by importing (and if desired changing) a net topology, importing the pre-trained weights of the model, setting the learning rate modifier parameter to 0 for the first few layers, and starting a training session with the desired data.

Training can also be performed on a different machine using Barista's \textit{remote sessions}. By specifying databases on different hosts and selecting one of them for training, models can be trained on other machines while locally monitoring the progress.

\section{Technical Details}
\label{sec:Characteristics}

The source code of Barista along with an online tutorial and user manual can be obtained at \baristaurl{} and is provided under MIT license to make the software as accessible to the research community as possible.  
The application is maintained by the Pattern Recognition and Image Analysis group at the Department of Computer Science at the University of M\"unster.
However, we appreciate any input from the community such as suggestions, comments, or patch submissions.
Barista is written in Python 2.7 using the Qt framework and PyQt5 for the graphical user interface as well as Seaborn\footnote{\url{https://seaborn.pydata.org/}} for plotting. 
For interaction with Caffe, the provided PyCaffe interface is used.
A complete list of dependencies can be found in the Barista documentation.
Although Barista runs on multiple operating systems such as Windows, MacOS, and various Linux distributions, the main development is currently targeting Ubuntu Linux.
\section{Conclusion and Future Work}
\label{sec:Conclusion}

We have presented Barista, an open-source graphical tool for designing and training deep neural networks using the popular Caffe deep learning framework.
Our application allows the graphical specification of the network topology and all of its parameters as well as the (hyper) parameters of the solver. 
The model can be trained by using Barista's input manager to specify training and validation data while the user can monitor the progress using the integrated plotting module. 
Our application supports different Caffe versions and provides a network interface to allow training on remote machines for more flexibility.

In the future, we plan to extend Barista in several directions. 
We would like to include more visualization options for the training as well as the weights of the trained model, such as the techniques proposed in~\citep{Zeiler14} or other methods from the field of information visualization.
Furthermore, we would like to provide configurable default parameter sets for frequently used layers and solver settings to further improve productivity when defining NNs and hyper parameters from scratch. 

\bibliographystyle{apalike}
\bibliography{bibliography}

\end{document}